\documentclass[10pt,twocolumn,letterpaper]{article}

\usepackage{iccv}
\usepackage{times}
\usepackage{epsfig}
\usepackage{graphicx}
\usepackage{amsmath}
\usepackage{amssymb}
\usepackage{booktabs}
\usepackage{wrapfig}

\usepackage[pagebackref=true,breaklinks=true,letterpaper=true,colorlinks,bookmarks=false]{hyperref}

\iccvfinalcopy 

\def\iccvPaperID{11370} 

\ificcvfinal\fi

\begin{document}

\title{Aggregation of disentanglement: Reconsidering domain variations in domain generalization}

\author{Daoan Zhang\textsuperscript{\rm 1}\footnotemark[1]\qquad~Mingkai Chen\textsuperscript{\rm 2}\footnotemark[1]\qquad~Chenming Li\textsuperscript{\rm 1}\qquad~Lingyun Huang\textsuperscript{\rm 3}\qquad~Jianguo Zhang\textsuperscript{\rm 1}\footnotetext{Equal contribution}\\
                {\textsuperscript{\rm 1} Southern University of Science and Technology}  \qquad
                {\textsuperscript{\rm 2} Stony Brook University} \\
                {\textsuperscript{\rm 3} PingAn Technology} \\
                }

\maketitle
\ificcvfinal\thispagestyle{empty}\fi

\def\thefootnote{*}\footnotetext{These authors contributed equally to this work}

\begin{abstract}
This research addresses the challenge of Domain Generalization (DG) in machine learning, which aims to improve the generalization of models across multiple domains. While previous methods focused on generating domain-invariant features, we propose a new approach that incorporates domain variations that contain useful classification-aware information for downstream tasks. Our proposed method, called Domain Disentanglement Network (DDN), decouples input images into \textbf{Domain Expert Features} and noise. The domain expert features are located in a learned latent space where images in each domain can be classified independently, allowing the implicit use of task-specific domain variations. In addition, we introduce a novel contrastive learning method to guide the domain expert features to form a more balanced and separable feature space. Our experimental results on widely used benchmarks, including PACS, VLCS, OfficeHome, DomainNet, and TerraIncognita, show that our method outperforms recently proposed alternatives, demonstrating its competitive performance.
\end{abstract}

\section{Introduction}

In the field of deep learning\cite{wang2022mvsnet, wang2023flora, wang2023ftso, zeng2022simple}, it is widely assumed that the training and test data conform to an independent and identical distribution\cite{wang2023Dionysus, zeng2023substructure}. However, in practical applications, it is often necessary for the models to perform well on out-of-distribution data, in addition to in-distribution data. This presents a formidable challenge for the models, as their ability to generalize is put to test. To address this challenge, domain generalization (DG) has been proposed as a crucial task for enhancing the model's generalization performance~\cite{muandet2013domain}.

The main idea behind DG is that during the training phase, only the source domains are accessible to the model, and the test data from target domains are not visible~\cite{ben2010theory,sabes1995advances}. Traditional DG methods aim to extract domain-invariant features across the source domains~\cite{arjovsky2019invariant,motiian2017unified,tzeng2014deep}. Specifically, the input data are transformed into a domain-invariant latent space, where the domain-invariant features can be effectively extracted. During inference, the test data from the unseen target domains are also projected onto the same latent space for downstream tasks such as classification. It is noteworthy that the domain-invariant features are those that are shared by \textit{all} source domains

\begin{figure}[t]
\setlength{\intextsep}{0pt}\
\centering
\includegraphics[width=0.9\columnwidth]{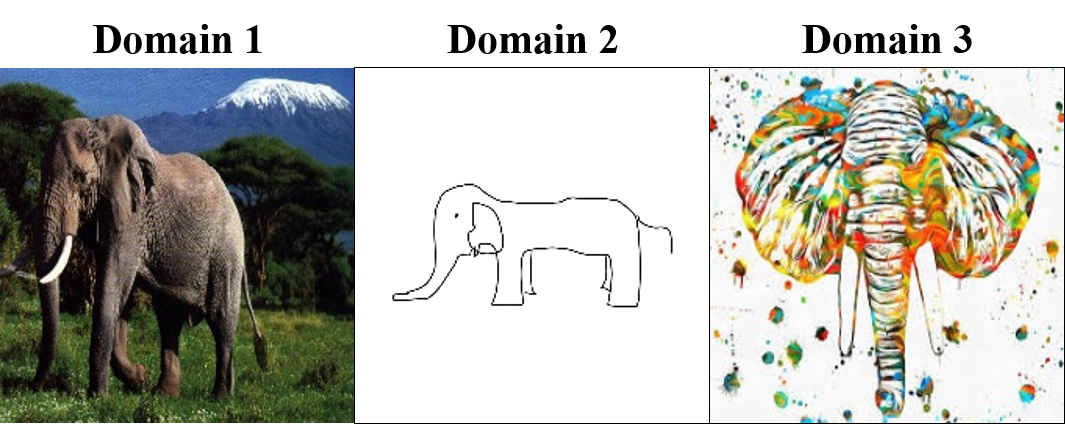} 
\caption{\textbf{Domains differ from each other.} Some features that benefit classification tasks in one domain have no presence in other domains. For example, skin texture features in domain 1 (left) help in identifying the elephant, and in domain 2 (middle), no texture features are present for classification, while in domain 3 (right), texture features impose lots of noise on the whole picture, making the classification tasks more challenging.}
\vspace{-1em}
\label{cont1}
\end{figure}

In this paper, we argue that such invariant features learned from source domains are not necessarily the best for classification, \ie, the best classification-aware features. Previous methods~\cite{bui2021exploiting,seo2020learning} have already introduced the domain specific features but they fail to consider whether the features is useful to downstream tasks. While we demonstrate the task-specific domain variations should have the following properties: (1) Domain variations should be specific for each domain, and (2) effective for downstream tasks. An illustrative example is presented in Fig.~\ref{cont1} to emphasize these properties. In domain 1, domain-specific features such as skin textures and surrounding context features play a crucial role in identifying an elephant, while in domain 2, these features are not available, and shape features are the principal cue. Domain 3 demonstrates a scenario in which texture features can have an adverse impact on elephant identification.

Moreover, from the view of unseen target domain, merely producing domain-invariant features from all source domains is inadequate for classification, because (1) the generated domain-invariant information from the source domains may not be suitable for the target domain and (2) domain variations in the target domain may also aid in classification.

Taking into account the task-specific domain variations, current domain generalization works can be classified into three categories: (a) data manipulation methods~\cite{volpi2018generalizing,xu2021fourier,zhou2021domain} that can diversify domain variations while maintaining domain-invariant features, (b) representation learning methods~\cite{ganin2016domain,nam2021reducing,seo2020learning} that aim to produce better domain-invariant representation from existing data, and (c) learning strategies~\cite{kim2021selfreg,shi2021gradient,yao2022pcl} that integrate other general machine learning paradigms into domain generalization.


Practically, we propose the adoption of task-specific domain variations in two key areas: 

(1) Disentanglement: At training time, for each source domain, the mapped features of the images should contain as much information as possible that is helpful for classification, \ie, the proposed features can be disentangled into the domain invariant and the task-specific domain variations.

(2) Aggregation: At the time of testing, the features mapped from the target test domains may also contain task-specific domain variations. While the task-specific domain variations from target domains can be seen as an aggregation (\eg, ideally linear combination) of the task-specific domain variations from training source domains.

To address these concerns, we introduce a new paradigm for DG classification tasks called the \textbf{Domain Disentanglement Network (DDN)}. To maintain task-specific domain variations for each domain, we design a special \textbf{Domain Expert Classifier (DEC)} for every domain as a domain specific ERM to reduce the domain specific noise. This classifier consists of a domain-specific classifier and a domain-specific projection head. Additionally, we develop a new contrastive approach known as \textbf{Domain Prototype Contrastive Learning (DPCL)} to disentangle domain invariant features and task-specific domain variations, respectively. This contrastive approach effectively elicits the discrepancies and similarities among source domains and maps features from the target domain to the aggregation of features from the source domains.

Specifically, our main contributions can be summarized as follows:

\begin{itemize}
    \item We propose a new perspective that emphasizes the importance of task-specific domain variations in domain generalization. We argue that to achieve effective generalization, all domain variations must be fully considered during training, and the task-specific domain variations from unseen target domains should be decoupled into the task-specific domain variations generated from accessible source domains. 
    
    \item We introduce an end-to-end paradigm for classification tasks in DG that consists of two parts. In the first part, images are sent into the domain expert classifier (DEC), which classifies the images and maintains task-specific domain variations. In the second part, then the domain prototype contrastive Learning (DPCL) is a reinforcement to learn task-specific domain variations from source domains.
    
    \item We demonstrate the effectiveness of our proposed approach by achieving state-of-the-art performance on multiple standard benchmarks, including PACS, VLCS, OfficeHome, TerraIncognita, and DomainNet. The experimental results validates the significance of our proposed perspective and the end-to-end paradigm for domain generalization.
\end{itemize}

\section{Related Work}

\subsection{Domain Generalization}

The domain generalization methods have been developed with the primary objective of learning a universal representation across all source domains. These methods have been broadly categorized into three groups: (1) Data-centered methods~\cite{huang2021fsdr,prakash2019structured,tobin2017domain}, which include data augmentation approaches aimed at presenting complex and diverse samples to the model. In addition, data generation techniques~\cite{li2021progressive,xu2021fourier} have been implemented that utilize generative methods to produce pseudo OOD images for further training. (2) Learning-strategy centered methods, which incorporate several strategies such as ensemble strategy~\cite{dubey2021adaptive,mancini2018best}, aimed at effectively combining multiple models to improve the generalization of the model, and gradient-manipulation methods~\cite{sun2016deep,tian2022neuron} that use gradient information to enhance the domain invariant representation. (3) Representation learning methods~\cite{bui2021exploiting,hu2020domain,jia2020single,peng2019moment}, which aim to generate domain invariant features that can minimize the representation discrepancy among domains, thus facilitating better generalization of the model. While we focus on the specific task-aware domain variations which is expected to enhance the overall classification accuracy and improve the model's performance in real-world scenarios.

\subsection{Contrastive Learning}

Contrastive learning~\cite{oord2018representation} is a common strategy to enhance the similarity between two instances in a positive pair and the dissimilarity in a negative pair. 

To achieve this, various contrastive learning algorithms have been proposed, such as SimCLR~\cite{chen2020simple}, which calculates the co-occurrence of multiple augmented views; MoCo~\cite{he2020momentum}, which utilizes a memory bank mechanism to handle a larger number of negative pairs during training; SimSiam~\cite{chen2021exploring} and BYOL~\cite{grill2020bootstrap}, which employ stop gradient operation to learn representations without a memory bank or a large batch size; DenseCL~\cite{wang2021dense}, utilizes dense alignments between semantic features; PixPro~\cite{xie2021propagate}, which leverages pixel-level pretext tasks and pixel-wise consistency to enhance downstream tasks; And PixContrast~\cite{wang2021exploring}, which expands contrastive learning to supervised learning. 

The domain generalization problem can also be considered as an dense contrastive learning problem. In dense prediction tasks such as segmentation, regional semantic features from different images are labeled with the same tag, while in the domain generalization problem, image features with different tags but in the same domain can be classified in the same domain category. Our proposed method introduces a prototype-based contrastive learning approach to better handle out-of-domain (OOD) tasks in domain generalization problems.

\section{Domain Variations in DG}

\subsection{Formulation}

We consider the classification task in DG as a mixture of minimizing the empirical risk of classification in each domain and measuring the decoupled weights of target domains over each source domain. We follow the leave-one-domain setting and assume that the source domains are $1,2,..., S$, and a single target domain is $T$, we then define $C_m$ as the domain invariant features, where $m$ is the $m^{th}$ class in a total of $M$ classes; $D_T$ as the task-specific domain variations over all the data in the target domain; $D_S$ as the task-specific domain variations over the source domains. From domain $s$, we sample $N$ data points which are defined as ${\{(x_i^s, y_i^s)\}}_{i=1}^N$.   

\subsection{Disentangle and Aggregate Domain Variations}

\begin{figure}[t]
\setlength{\belowcaptionskip}{-10pt}
\centering
\includegraphics[width=0.8\columnwidth]{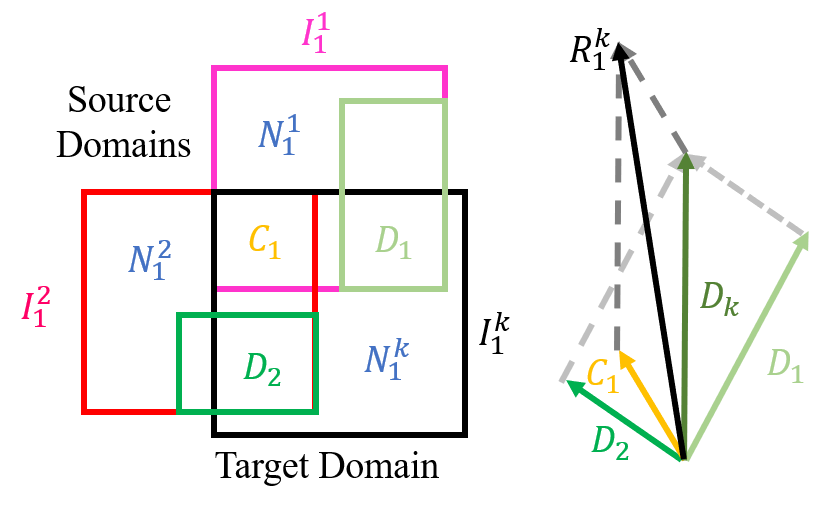} 
\caption{\textbf{Left:} The relationships between source and target domains.; \textbf{Right:} The disentangled task-specific domain variations and domain expert features of target domains.}
\label{venn}
\end{figure}

Taking the classification as an example, we first conceptually decompose the generated information from an image $I^s_m$ from domain $s$, class $m$ into three component vectors:
\begin{equation}
    I^s_m = C_m + D_s + N^s_m,
    \label{000}
\end{equation}
where $C_m$ denotes the domain invariant features, $D_s$ denotes the task-specific domain variations, and $N^n_m$ denotes the useless information for classification.

The Venn diagram depicted in Fig.~\ref{venn} (left) provides a visual representation of our conceptual decomposition. Specifically, $I^1_1$ and $I^2_1$ are instances of class 1 sampled from domains 1 and 2, respectively. As such, they share the same domain invariant features $C_1$. However, $I^1_1$ and $I^2_1$ also contain task-specific domain variations $D_1$ and $D_2$, respectively, as well as noisy information $N_1^1$ and $N_1^2$. In the roght of the graph, for an image $I_1^k$ sampled from an unseen target domain $k$, the task-specific domain variations can be expressed as a linear combination of $D_1$ and $D_2$, which is an ideal scenario. With this hypothesis, we introduce the \textbf{Domain Expert Features} $R^s_m$, depicted in Fig.~\ref{venn} (right), which combine domain invariant features with task-specific domain variations, and are deemed sufficient for classification purposes. The feature is defined as:

\begin{equation}
\label{equ-1}
R^s_m = C_m + D_s,
\end{equation}

In training, we propose to generate domain expert features for each source domain. In inference, as the unseen target domain is hard to analyze for the trained model, we propose to use the existing task-specific domain variations to represent the task-specific domain variations from the target domain which can be seen as an aggregation of the disentangled domain expert features. We first denote the relationship between the task-specific domain variations from source domains and the target domain as:
\begin{equation}
\label{eq1000}
    D_T = \sum_{s=1}^{S}{w_{Ts}D_{s}},
\end{equation}
Then,
\begin{equation}
\label{eq0}
    C_m + D_T = C_m + \sum_{s=1}^{S}{w_{Ts}D_{s}},
\end{equation}
where $w_{T} = [w_{T1}, ..., w_{Ts}]$ is the domain-specific weights of decoupled target task-specific domains variations over the source task-specific domain variations, which lies on $w_{Ts} \in \mathbb{R}, {\left\| w_{T} \right\|}_1 = 1, w_{Ts} \geq 0$. Thus, according to Eq.~\ref{equ-1}, Eq.~\ref{eq0} can be modified to:
\begin{equation}
\label{eq1}
    R_m^T = \sum_{s=1}^{S}{w_{Ts}}R_m^s,
\end{equation}
Thus, $R^T = \sum_{m=1}^{M} {R^T_m} = \sum_{m=1}^{M}\sum_{s=1}^{S}R_m^s = \sum_{s=1}^{S}R^s$.

As the key point of DG problems is to reduce the discrepancy between the source domains features and target domain features~\cite{volpi2018generalizing}, which can be written as:
\begin{equation}
\label{eq3}
    \mathcal{L_P}(E(x;\theta_E); \theta_P) = \frac{1}{S}\sum_{s=1}^{S}{\textit{d}(R^s, R^T)},
\end{equation}
where $E(\cdot; \theta_E)$ is defined as the backbone of the model, $\mathcal{L_P}$ is the loss of domain awareness head, $\textit{d}(\cdot, \cdot)$ is a distance to measure the discrepancy between the features. According to our assumption, the domain expert features $R^s$ are diverse from each other, but how to map the features into an appropriate latent space that can better demonstrate target features $R^T$ tends to be explored.   

To better estimate the discrepancy, following~\cite{arjovsky2017wasserstein}, we utilize the commonly used metric for distributions: Wasserstein-1 distance as the measurement. Thus, Eq.~\ref{eq3} can be written as:
\begin{equation}
\label{eq4}
  \begin{aligned}
    & \mathcal{L_P}(E(x;\theta_E); \theta_P) \\
    &= \frac{1}{S} \sum_{s=1}^{S} \left(\underset{x\sim R^s}{\mathbb{E}} [f_s(x)] -\underset{x\sim R^T}{\mathbb{E}}[f_s(x)] \right), \\
  \end{aligned}
\end{equation}
where $f_s(\cdot)$ is the 1-Lipschitz function $f: X \to \mathbb{R}$ which can be implemented by a neural network~\cite{arjovsky2017wasserstein} for every source domain.

We introduce Eq.~\ref{eq1} to the penalty Eq.~\ref{eq4}, it turns to:
\begin{equation}
\label{eq5}
  \begin{aligned}
    & \mathcal{L_P}(E(x;\theta_E); \theta_P) = \frac{1}{S} \sum_{s=1}^{S} \left(\underset{x\sim R_s}{\mathbb{E}}[\gamma_s^T F(E(x))]\right) 
  \end{aligned}
\end{equation}

Where $F(E(x)) = {[f_1(E(x)),..., f_S(E(x))]}^T$ is the concatenation of the 1-Lipschitz functions for every source domain, and $\gamma_s^T$ is denoted as:
\begin{equation}
\label{eq6}
\gamma_s=\left\{
\begin{aligned}
1-w_{ss^{'}}, &  & s=s^{'} \\
-w_{ss^{'}}, &  & s \neq s^{'} 
\end{aligned}
\right.
\end{equation}

where $s^{'} = 1,..., S$, $w_{ss^{'}}$ is the weight of the feature from $s$ over the feature from $s^{'}$. 

As we assume that $N$ cases are sampled from each source domain, the Eq.~\ref{eq5} turns out to be:
\begin{equation}
\label{eq7}
  \begin{aligned}
    & \mathcal{L_P}(E(x;\theta_E); \theta_P) \simeq  \frac{1}{SN} \sum_{s=1}^{S} \left[ \sum_{n=1}^{N} {\gamma_s^T} F(E(x))
\right]
  \end{aligned}
\end{equation}

Notice that, we have $w_{ss^{'}} \in [0, 1]$, and according to Eq.~\ref{eq6} and~\ref{eq7}, to minimize the loss term $\mathcal{L_P}(E(x;\theta_E); \theta_P)$, for every data from each domain, when $s=s^{'}$, the function $f_s(E(x))$ should be minimized, and when $s \neq s^{'}$, the function $f_s(E(x))$ should be maximized. This means the distance between features from different domains should be pulled away, and the distance between features from the same domain should be narrowed.
Practically, we design the Domain Prototype Contrastive Learning to modify the distances.  

\section{Domain Disentanglement Network (DDN)} 

\begin{figure*}[t]
\setlength{\belowcaptionskip}{-10pt}
\centering
\includegraphics[width=0.83\textwidth]{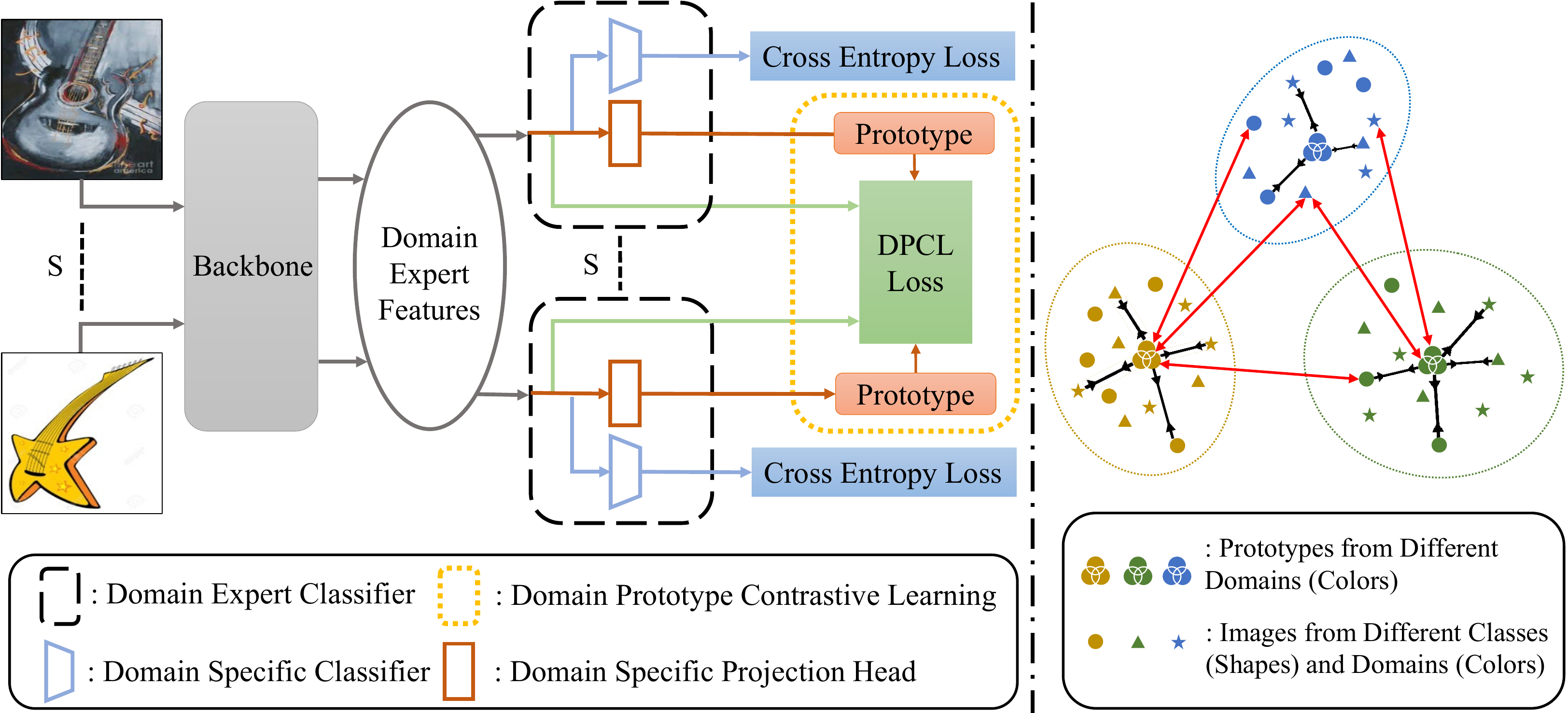} 
\caption{\textbf{Right:} DDN Network. Image features from different domains are sent into different domain expert classifiers based on domain labels. The relations between the domain expert features are calculated through the Domain Prototype Contrastive Learning strategy. \textbf{Left:} Schematic Diagram of Domain Prototype Contrastive Learning. Black lines mean pushing in-domain features near the prototype of the domain, and red lines mean pushing features from other domains far away from the prototype of the domain. }
\label{fig2}
\end{figure*}
To satisfy the requirements, we designed a novel network called \textbf{Domain Disentanglement Network (DDN)}. Compared to traditional methods, the network considers and decouples both domain invariant and task-specific domain variations for classification when in training (Fig.~\ref{fig2}. left), and in the inference period, classification-aware features from unseen target domain are identified and aggregated according to learned source domains (Fig.~\ref{test}) and are classified through the corresponding classifiers.
 
\subsection{Domain Prototype Contrastive Learning (DPCL)}

\textbf{Motivations.} Based on Eq.~\ref{eq7}, we propound a newly designed contrastive learning-based method called Domain Prototype Contrastive Learning (DPCL).
This method aims to modify the discrepancies between domain expert features when training and decouple domain expert features from the target domain when inference (see Fig.~\ref{fig2}. right). We notice that in the inference, the model has to calculate the aggregation weights over domain expert features of each source domain for target test features. The commonly used paradigm, memory bank~\cite{he2020momentum}, can store the domain expert features from source domains. However, due to the discrepancies of different classes, features from the same domain can be various from each other, and the stored memory is hard to converge. Therefore, the memory bank mechanism fails to learn sufficient memory on DG tasks. To relieve the problem, instead of storing the domain specific features, we proposed a domain specific projection head to build the prototype for calculating the weights.

\textbf{Whole Structure.} In our method, we generate the prototype~\textit{dynamically}, and we use a tiny network to aggregate prototypes separately for each domain in the batch (see Fig.~\ref{fig2}. left). This can relieve the discrepancy between the features as well as converge faster. 

Thus, when we sample $N$ cases from one source domain $s^+$, our DCPL loss is designed as follows:
\begin{equation}
\label{dpcl}
    \begin{aligned}
    & \mathcal{L_P}(E(x;\theta_E); \theta_P, s^+) \\
    & = -log \frac{exp[{ \sum^{N}_{n=1}cos(E(x^{s^+}_n, q^{s^+}))}]}{\sum_{s=1}^{S}exp[{\sum_{n=1}^{N}}cos(E(x^{s}_n), q^{s^+})]}
    \end{aligned}
\end{equation}
where $cos(\cdot, \cdot)$ denotes cosine similarities between two embeddings. $x^{s^+}_n$ denotes the $n^{th}$ case sampled from the chosen $s^+$ domain, and $q^{s^+}$ is designed as:
\begin{equation}
    q^{s^+} = \frac{1}{n} \sum_{n=1}^{N} P^{s^+}(E(x^{s{^+}}_n))
\end{equation}
where $P^{s^+}$ denotes the projection head for domain $s^+$.
Thus, with the DPCL module and the domain expert classifier, When training with source domains, both the classification-aware domain invariant and variant features can be disentangled from the source domains for more precise classification. When in the inference, the test domain features can be represented as the aggregated domain expert features from source domains to better understand the unseen domains, 

\textbf{Difference between DPCL and existing contrastive DG paradigms.}

\begin{figure}[t]
\centering
\includegraphics[width=0.8\columnwidth]{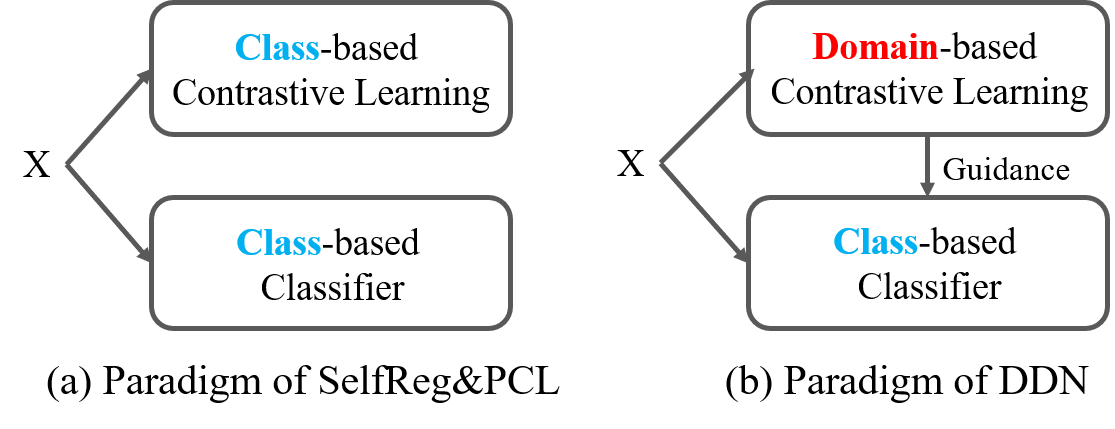} 
\caption{The sketch map of DDN and traditional contrastive learning methods in DG}
\label{contrast}
\end{figure}

The key differences can be concluded as follows:

(1) Different solutions for contrastive learning: Traditional contrastive paradigms like SelfReg~\cite{kim2021selfreg} and PCL~\cite{yao2022pcl} are \textit{classification-based} contrastive learning while we propose a \textit{domain-based} prototype contrastive learning paradigm. As shown in Fig.~\ref{contrast}, classification-based contrastive learning tries to map a uniform latent space for classification but fails to consider the information from domains. Also, in contrastive learning, too difficult sample pairs from different domains can hamper the model generalization~\cite{tian2020makes}. In our proposed domain-based prototype contrastive learning, the domain prototype provides domain-specific information for the classifier, and we can avoid facing the problem of difficult samples.

(2) Different purposes of using contrastive learning: Conventional contrastive paradigm aims to obtain better domain invariant features to generalize the model. However, we focus on rather than domain invariant features but to generate more uniform representations of task-specific domain variations to better represent the domain information for unknown test domains.

\subsection{Domain Expert Classifier (DEC)}
We design a specific classifier DEC for every input domain. Each classifier consists of a domain-aware classifier as a ERM model, a project head for generating prototypes, and a skip connection for DPCL to calculate distance. When we take a look at Eq.~\ref{000}, the noise $N^s_m$ in each domain can be removed by training the per-domain ERM classifiers. Thus, for each per-domain ERM classifier, the input domain expert features can be seen as the $C_m + D_s$ in Eq.~\ref{000}. With the combination of all the per-domain ERM classifier, the backbone can generate all the domain expert features for different domains. Thus, the total classification loss is:
\begin{equation}
    \mathcal{L}_Y(x, y) = \frac{1}{SNM} \sum_{s=1}^{S}\sum_{n=1}^{N}\sum_{m=1}^{M} y_{n,m}^{s} log Y^s_m [E(x^s_{n})]
\end{equation}

where $Y^s_m$ denotes the $m^{th}$ entry of the output of domain $s$.

The structure of the domain expert classifier is proposed in Fig.~\ref{fig2}, left. We use several MLP layers as classifiers and project heads.

\subsection{Overall Strategy}
Thus, the loss in our methods is designed to be:
\begin{equation}
    \mathcal{L}(x, y, t) = \mathcal{L}_Y(x, y) + \lambda \mathcal{L_P}(x, t) 
\end{equation}
where $x$ is the input image, $y$ is the classification ground truth label, and $t$ is the domain ground truth label. $\lambda$ is the loss weight. In the training period, all domain-specific classifiers are trained simultaneously, and the domain prototype contrastive learning is also processed to maintain the classification-aware features. 

\begin{figure}[t]
\centering
\includegraphics[width=0.9\columnwidth]{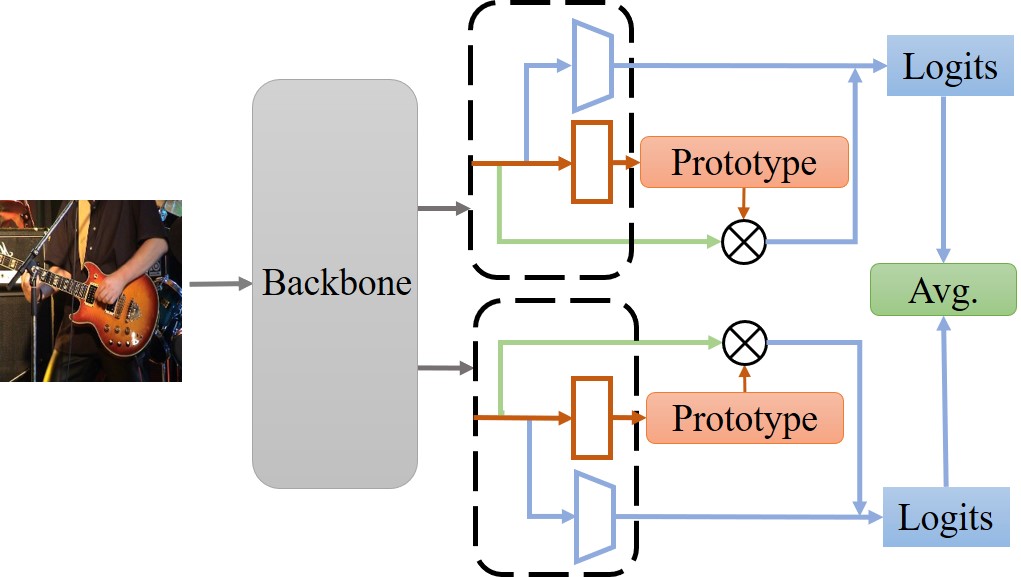} 
\caption{\textbf{The sketch map of inference.} Images are sent into the backbone to achieve the domain expert features, and then the features are fed into each domain expert classifier. The classifier weights are calculated via the distance between the prototype and the features. Afterward, the result can be obtained by aggregating the weighted result of each classifier.}
\label{test}
\end{figure}

In the inference period, as shown in Fig.~\ref{test}, the domain expert features generated from the target domain are sent into all domain expert classifiers, and the prototypes from each source domain are calculated for measuring the weights for each domain expert classifier. Then the weights are imposed on the classification result of each classifier respectively. In the end, we calculate the average result of all the classifiers to be the final decision.

\section{Experiments}

\subsection{Implementation Details}

\textbf{Datasets.}
To evaluate our proposed methods, we choose five larger datasets in DomainBed~\cite{gulrajani2020search}: PACS~\cite{li2017deeper}, VLCS~\cite{torralba2011unbiased}, OfficeHome~\cite{venkateswara2017deep}, DomainNet~\cite{peng2019moment} and TerraIncognita~\cite{beery2018recognition}.
The PACS dataset consists of 10k images from four different domains. VLCS dataset consists of four domains and five classes. OfficeHome dataset is sampled from four domains with around 15500 images of 65 categories. DomainNet contains 0.6 million images of 345 classes sampled from six domains. TerraIncognita consists of 24788 images, ten classes, and four domains.

\textbf{Baselines.}
Our experiments are based on the open-source code SWAD~\cite{cha2021swad}, and DomainBed~\cite{gulrajani2020search} benchmarks. Other training strategies and data augmentation we use remains the same with SWAD~\cite{cha2021swad} and DomainBed~\cite{gulrajani2020search}. Follow~\cite{gulrajani2020search}, for PACS, VLCS, OfficeHome, and TerraIncognita; we train our model for 5000 iterations. For DomainNet, we train our model for 15000 iterations. We use the leave-one-out setting, \ie; a specific single domain is used as a test domain and the others as a training domain.  

\textbf{Hyperparameter search and model selection.}
For DomainBed baseline, we conduct a random search of 20 trials over the hyperparameter distribution for each algorithm and test domain. Results are averaged over three seeds. We set the search distribution of hyparameter $\lambda$ as $\{1, 5, 10, 20, 30\}$ . Specifically, the model selection strategy we utilize is the 'train-domain validation (oracle)': the validation set follows the same distribution as the test domain. For the SWAD baseline, we follow the initial settings in SWAD~\cite{cha2021swad} to implement the experiments, and the model selection is not required.

\textbf{Model architectures.}
The pretrained model we used is Resnet-50 which is pretrained on ImageNet. We do not use any additional information or training loops compared to those who use GAN or complex data augmentations. Besides the Resnet-50 backbone, the proposed classifier architecture is a one-layer MLP, and the projection head is a two-layer MLP. We used cosine similarity loss to calculate the distance in DPCL.

\subsection{Qualitative Results}
We compared our methods with current state-of-the-art methods. For a fair comparison, all methods utilize Resnet-50 pre-trained on ImageNet. The first set of comparative experiments is based on DomainBed~\cite{gulrajani2020search} as is shown in Table.~\ref{tab:ALL}, both results in most single domains, and on average of our model prove that our model can effectively generalize on unseen domains. Notice that, Our methods do not use any additional training loop like GAN or adversarial training, which is not that feasible in practice. Additionally, our method achieves better results compared to the previous methods~\cite {kim2021selfreg},~\cite{wang2020learning}, which are based on contrastive learning. 

The second set of comparative experiments is based on SWAD~\cite{cha2021swad}, which is presented in Table.~\ref{swad1}. SWAD is a solid baseline that can be concatenated with several DG methods. In the table, we compare our method to the current state-of-the-art contrastive learning based method PCL~\cite{yao2022pcl} on PACS and VLCS datasets. Our method achieves over a 1\% increase on both of the datasets. 

Moreover, we observe that our method outperforms other methods on two kinds of target domains: The first kind of domains is the domains with adequate task-specific domain variations, like the SUN09 (S) and VOC2007 (V) domains which are sampled from the real-world in VLCS dataset. The second is the kinds of domains with little or even no task-specific domain variations, like the Arts (A) in PACS and Sketch (S) in PACS. While for those domains that have unmatched task-specific domain variations, \eg, background information, to the semantic objectives, like Cartoon (C) in PACS and Caltech101 (C) in VLCS, our methods fail to achieve the best performance but still achieve competitive performance. These phenomenons indicate that, when our model predicts an image, it not only considers the domain invariant information but also considers task-specific domain variations. That is why our model can outperform many complex models that only focus on domain invariant information.

\begin{table*}[]
    \centering
    \begin{tabular}{l|cccccc}
    \hline
    Algorithm  & VLCS & PACS & OfficeHome & TerraInc & DomainNet & Average \\
    \toprule
    ERM~\cite{vapnik1999overview}            & 77.6 $\pm$ 0.3            & 86.7 $\pm$ 0.3            & 66.4 $\pm$ 0.5            & {53.0} $\pm$ 0.3            & 41.3 $\pm$ 0.1            & 65.0                      \\
    
    IRM~\cite{arjovsky2019invariant}                 & 76.9 $\pm$ 0.6            & 84.5 $\pm$ 1.1            & 63.0 $\pm$ 2.7            & 50.5 $\pm$ 0.7            & 28.0 $\pm$ 5.1            & 60.6                      \\
    
    GroupDRO~\cite{sagawa2019distributionally}               & 77.4 $\pm$ 0.5            & 87.1 $\pm$ 0.1            & 66.2 $\pm$ 0.6            & 52.4 $\pm$ 0.1            & 33.4 $\pm$ 0.3            & 63.3                      \\
    
    Mixup~\cite{yan2020improve}                & 78.1 $\pm$ 0.3            & 86.8 $\pm$ 0.3            & 68.0 $\pm$ 0.2            & \textbf{54.4} $\pm$ 0.3            & 39.6 $\pm$ 0.1            & 65.4                     \\
    
    MLDG~\cite{li2018learning}              & 77.5 $\pm$ 0.1            & 86.8 $\pm$ 0.4            & 66.6 $\pm$ 0.3            & 52.0 $\pm$ 0.1            & 41.6 $\pm$ 0.1            & 64.9                      \\
    
    CORAL~\cite{sun2016deep}               & 77.7 $\pm$ 0.2            & 87.1 $\pm$ 0.5            & 68.4 $\pm$ 0.2            & 52.8 $\pm$ 0.2            & 41.8 $\pm$ 0.1            & 65.6                      \\
    
    MMD~\cite{li2018domain}              & 77.9 $\pm$ 0.1            & 87.2 $\pm$ 0.1            & 66.2 $\pm$ 0.3            & 52.0 $\pm$ 0.4            & 23.5 $\pm$ 9.4            & 61.4                      \\
    
    DANN~\cite{ganin2016domain}        & 79.7 $\pm$ 0.5            & 85.2 $\pm$ 0.2            & 65.3 $\pm$ 0.8            & 50.6 $\pm$ 0.4            & 38.3 $\pm$ 0.1            & 63.8                      \\
    
    CDANN~\cite{li2018deep}         & 79.9 $\pm$ 0.2            & 85.8 $\pm$ 0.8            & 65.3 $\pm$ 0.5            & 50.8 $\pm$ 0.6            & 38.5 $\pm$ 0.2            & 64.1                      \\
    
    MTL~\cite{blanchard2021domain}               & 77.7 $\pm$ 0.5            & 86.7 $\pm$ 0.2            & 66.5 $\pm$ 0.4            & 52.2 $\pm$ 0.4            & 40.8 $\pm$ 0.1            & 64.8                     \\
    
    SagNet~\cite{nam2021reducing}          & 77.6 $\pm$ 0.1            & 86.4 $\pm$ 0.4            & 67.5 $\pm$ 0.2            & 52.5 $\pm$ 0.4            & 40.8 $\pm$ 0.2            & 65.0                      \\
    
    ARM~\cite{zhang2021adaptive}              & 77.8 $\pm$ 0.3            & 85.8 $\pm$ 0.2            & 64.8 $\pm$ 0.4            & 51.2 $\pm$ 0.5            & 36.0 $\pm$ 0.2            & 63.1                      \\
    
    VREx~\cite{krueger2021out}        & 78.1 $\pm$ 0.2            & 87.2 $\pm$ 0.6            & 65.7 $\pm$ 0.3            & 51.4 $\pm$ 0.5            & 30.1 $\pm$ 3.7            & 62.5                      \\
    
    RSC~\cite{huang2020self}           & 77.8 $\pm$ 0.6            & 86.2 $\pm$ 0.5            & 66.5 $\pm$ 0.6            & 52.1 $\pm$ 0.2            & 38.9 $\pm$ 0.6            & 64.3                      \\
    
    \midrule
    SelfReg~\cite{kim2021selfreg} &  77.5 $\pm$ 0.0 & 86.5 $\pm$ 0.3 & \underline{69.4} $\pm$ 0.2 & {51.0} $\pm$ 0.4 &  \textbf{44.6} $\pm$ 0.1 & \underline{65.8}\\
    DA-CORAL~\cite{dubey2021adaptive} & 78.5 $\pm$ 0.4 & 84.5 $\pm$ 0.7 & 68.9 $\pm$ 0.4 & 48.1 $\pm$ 0.3 & \underline{43.9} $\pm$ 0.3 & 64.7 \\ 
    mDSDI~\cite{bui2021exploiting} & \underline{79.0} $\pm$ 0.3 & 86.2 $\pm$ 0.2 & 69.2 $\pm$ 0.4 & 48.1 $\pm$ 1.4 & 42.8 $\pm$ 0.1 & 65.0\\
    ITL-Net~\cite{gao2022loss} & 78.9 $\pm$ 0.7 & 86.4 $\pm$ 0.5 & 69.3 $\pm$ 0.4 & {51.0} $\pm$ 2.1 & 41.6 $\pm$ 0.3 & 65.5 \\
    Fishr~\cite{rame2022fishr} & 78.2 $\pm$ 0.2 & {86.9} $\pm$ 0.2 & 68.2 $\pm$ 0.2 &  \underline{53.6} $\pm$ 0.4 & 41.8 $\pm$ 0.2 & {65.7}  \\
    \midrule
    DDN & \textbf{80.3} $\pm$ 0.2 &  \textbf{87.2} $\pm$ 0.2 &  \textbf{69.5} $\pm$ 0.2 & 50.4 $\pm$ 0.5 & {43.4} $\pm$ 0.1 & \textbf{66.2} \\
    \bottomrule
    \end{tabular}
    \caption{DomainBed accuracies (\%). We utilize DomainBed~\cite{gulrajani2020search} as the backbone. We average over 3 seeds and report standard deviations. }
    \label{tab:ALL}
\end{table*}

\begin{table*}[]
    \setlength{\belowcaptionskip}{-10pt}
    \centering
    \begin{tabular}{l|ccccc|ccccc}
    \toprule
    Algorithm  & A & C & P & S & Avg. & C & L & S & V & Avg. \\
    \hline
    SWAD~\cite{cha2021swad} & 89.3 & 83.4 & 97.3 & 82.5 & 88.1 & 98.4 & \textbf{63.6} & 72.0 & 74.5 & 77.1 \\
    SWAD + PCL~\cite{yao2022pcl}   & 90.2 & \textbf{83.9} & \textbf{98.1} & 82.6 & 88.7 & \textbf{99.0} & \textbf{63.6} & 73.8 & 75.6 & 78.0  \\
    \midrule
    SWAD + DDN & \textbf{90.0} & 83.2 & 97.9 & \textbf{84.0} & \textbf{88.8} & 97.9 & 62.9 & \textbf{80.1} & \textbf{78.0} & \textbf{79.7} \\
    \bottomrule
    \end{tabular}
    \caption{SWAD accuracies (\%). We also utilize SWAD~\cite{cha2021swad} as the backbone on PACS and VLCS datasets.}
    \label{swad1}
\end{table*}

\subsection{Visualization Results}
We visualize the t-SNE results on several related works and our method. Notice that the visualization is based on the training set of the PACS dataset. We argue that in DG tasks, t-SNE presented on the unseen test dataset has artificial bias due to the selection of the dataset. Thus we focus on the uniformity and alignment measurement in self-supervised learning mentioned in~\cite{wang2020understanding}, as both DG and self-supervised model need to generalize to various domains. In~\cite{wang2020understanding}, the authors argue that (1) alignment (closeness) of features from positive pairs and (2) uniformity of the induced distribution of the features on the hypersphere can be seen as a measurement of representation quality.

\begin{figure}[t]
\centering
\includegraphics[width=0.9\columnwidth]{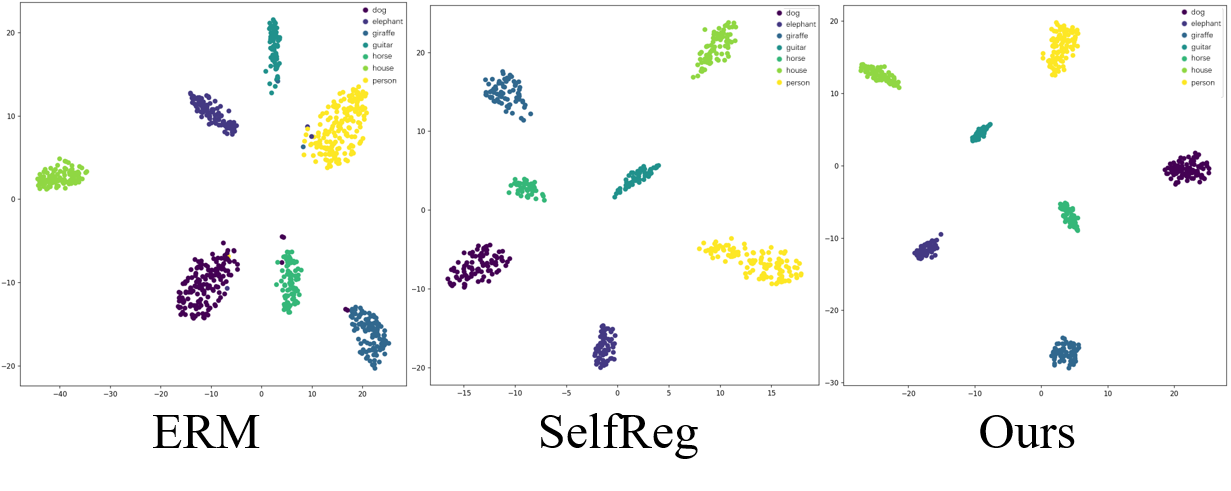} 
\caption{From left to right, we propose the t-SNE visualizations of  ERM~\cite{vapnik1999overview}, SelfReg~\cite{kim2021selfreg}, and our method. Our approach is tighter within classes and more uniform across classes.}
\vspace{-10pt}
\label{tsne}
\end{figure}

As shown in Fig.~\ref{tsne}, from the perspective of alignment, our method is much tighter within class compared to ERM~\cite{vapnik1999overview}, and SelfReg~\cite{kim2021selfreg}. Also, The uniformity of SelfReg and our method surpass the ERM methods, which is why SelfReg performs better than ERM and worse than ours.
The t-SNE results prove that our domain expert features can aggregate compact clusters simultaneously and hold a uniform distribution across source domains. Therefore, our model can achieve a good classification result while proficiently mapping the features from target domains to the domain expert space formed by the known source domains.

\subsection{Discrepancies among the Domains}

\begin{figure}[t]
\setlength{\belowcaptionskip}{-10pt}
\centering
\includegraphics[width=0.8\columnwidth]{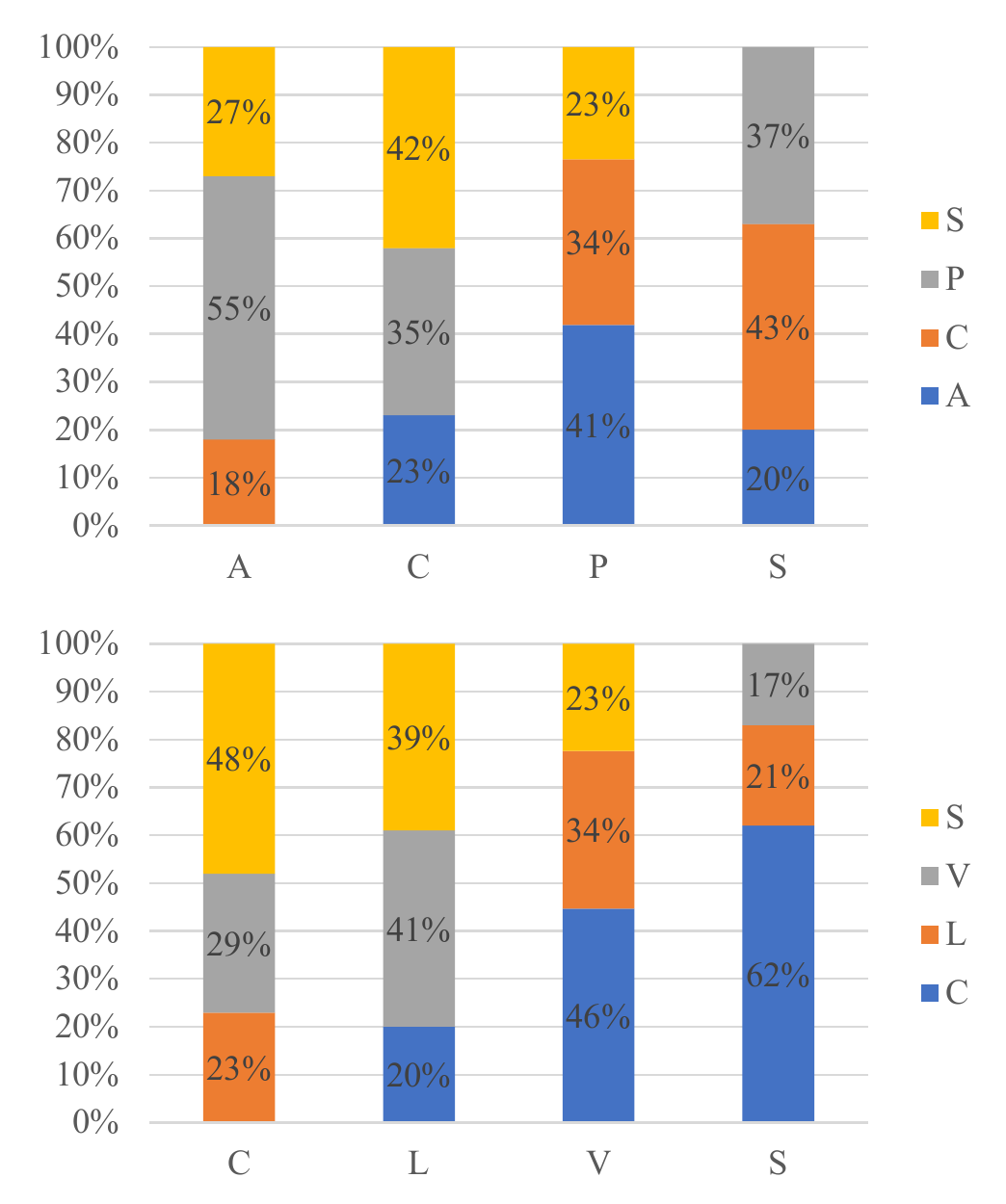} 
\caption{Discrepancies between domains on PACS and VLCS.}
\label{cont}
\end{figure}

In our methods, we disentangle the features from the source domains and aggregate the domain expert features for unseen domains. The natural distribution divergence can be measured by counting the weights among each domain. The divergence is presented in Fig.~\ref{cont}. We randomly choose 128 cases from target test domains and utilize the best model to evaluate the images and access the weights. 

For PACS, Arts (A) and Sketch (S) overlap greatly as both lack rich background (can be seen as the main part of task-specific domain variations) information, and our model may focus on the shape prior for classification. For VLCS, Caltech101 (C) and SUN97 (S) remain a high overlap. When the test domain is C, the model has a bigger chance to count on the classifier of S. However, the background of C is always irrelevant to the semantic objects, which may mislead the classifier to a wrong classification. Conversely, When the test domain is S, the model has a bigger chance to count on the classifier of C. As task-specific domain variations in C are misleading, the C classifier would focus on the foreground. Thus, the test S domain can achieve an impressive result.

\subsection{Ablation Study}

\textbf{Ablation study on DPCL.} 
We first analyze the effectiveness of the domain prototype contrastive learning strategy. When we mute the strategy, the model turns out to be an ensemble learning form. We utilize proxy-anchor loss~\cite{kim2020proxy} and memory bank mechanism~\cite{he2020momentum} to test the validity of our method. The result is shown in Table.~\ref{ab1}. The result proves that domain prototype contrastive learning is an indispensable module. 

\begin{table}[]
    \setlength{\belowcaptionskip}{-20pt}
    \centering
    \begin{tabular}{l|ccccl}
    \toprule
    Algorithm  & A & C & P & S & Avg. \\
    \hline
    w/o DPCL    & 87.3 & \textbf{81.1} & \textbf{97.2} & 82.4 & 87.0 \\
    + proxy-anchor & 86.4 & 79.9 &96.7 & 81.6& 86.2\\
    + memory bank & 82.3& 74.8& 93.2& 76.9& 81.8\\
    \midrule
    w/o shared classifier    & 85.3 & 74.9 & 96.0 & 79.7 & 84.0 \\
    \midrule
    Ours & \textbf{87.4} & 80.6 & 97.1 & \textbf{83.8} & \textbf{87.2} \\
    \bottomrule
    \end{tabular}
    \caption{Ablation study on domain prototype contrastive learning strategy and domain expert classifiers. The performance is much worse if we drop or change these main components.}
    \label{ab1}
\end{table}

\textbf{Ablation study on shared domain expert classifiers.}
Each domain should have an independent classifier, thus can better fit the matched domain expert features. We conduct the ablation study on the separated and shared classifier. The results are shown in Table.~\ref{ab1} proves that the shared classifier failed to be an expert classifier for all the domains. Therefore, the use of a separated classifier can better generate classification-aware features.

\textbf{Ablation study on batch size.}

\begin{wrapfigure}{}{0.45\linewidth}
 	\centering
 	\includegraphics[width=1\linewidth]{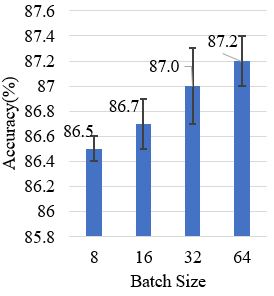} 
 	\caption{Ablation study of batch size on PACS.}
 	\label{bs}
\end{wrapfigure}

Due to the sensitivity of contrastive learning to batch size, we also ablate the batch size. The result is presented in Fig.~\ref{bs}. With the increase in batch size from 8 to 64, we can see that the performance proliferates.

\section{Limitations}
Although we present a fundamental model for the DG problem and achieve impressive results, there is an irreparable gap between target and source domains. It is still challenging to disentangle features for some far different domains from known source domains. However, this concern can be further relieved by adding more domains in training to let the model \textit{see} more task-specific domain variations.

\section{Conclusion}
We proposed a new perspective to utilize and disentangle domain expert features which consist of domain invariant and task-specific domain variations. In inference, our model effectively aggregates the knowledge from known domains to represent the target domain. We build a novel and robust network for the classification task in DG. Specially, we designed a contrastive learning based paradigm to address task-specific domain variations from the test domain. Our method outperforms the current methods and can improve further if there are more source domains.

{\small
\bibliographystyle{ieee_fullname}
\bibliography{egbib}
}

\end{document}